# Evaluating software-based fingerprint liveness detection using Convolutional Networks and Local Binary Patterns

Rodrigo Frassetto Nogueira, Roberto de Alencar Lotufo
Department of Control and Automation (DCA)
University of Campinas (UNICAMP)
Campinas, Brazil
rodrigonogueira4@gmail.com; lotufo@unicamp.br

Rubens Campos Machado
Centro de Tecnologia da Informação Renato Archer (CTI)
Campinas, Brazil
rubens.campos.machado@gmail.com

*Abstract*—With the growing use of biometric authentication systems in the past years, spoof fingerprint detection has become increasingly important. In this work, we implement and evaluate two different feature extraction techniques for software-based fingerprint liveness detection: Convolutional Networks with random weights and Local Binary Patterns. Both techniques were used in conjunction with a Support Vector Machine (SVM) classifier. Dataset Augmentation was used to increase classifier's performance and a variety of preprocessing operations were tested, such as frequency filtering, contrast equalization, and region of interest filtering. The experiments were made on the datasets used in The Liveness Detection Competition of years 2009, 2011 and 2013, which comprise almost 50,000 real and fake fingerprints' images. Our best method achieves an overall rate of 95.2% of correctly classified samples - an improvement of 35% in test error when compared with the best previously published results.

## I. INTRODUCTION

The basic aim of biometrics is to automatically discriminate subjects in a reliable way and according to some target application based on one or more signals derived from physical or behavioral traits, such as fingerprint, face, iris, voice, hand, or written signature. Biometric technology presents several advantages over classical security methods based on either some information (PIN, Password, etc.) or physical devices (key, card, etc.) [1]. However, providing to the sensor a fake physical biometric can be an easy way to overtake the system's security. Fingerprints, in particular, can be easily spoofed from common materials, such as gelatin, silicone, and wood glue [2]. Therefore, a safe fingerprint system must distinguish correctly a spoof from an authentic finger.

Different fingerprint liveness detection algorithms have been proposed [3] [4] [5], and they can be broadly divided into two approaches: Hardware and Software. In the hardware approach a specific device is added to the sensor in order to detect particular properties of a living trait such as the blood pressure [6], skin distortion [7], or the odor [8]. In the software approach, which is used in this work, fake traits are detected once the sample has been acquired with a standard sensor.

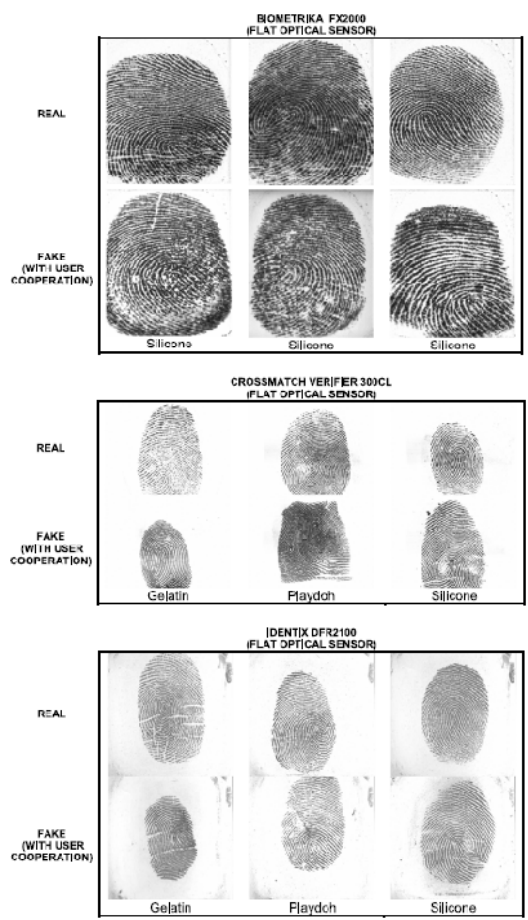

Fig. 1  Typical examples of real and fake fingerprint images that can be obtained from the LivDet2009 database used in the experiments. Figure extracted from **[9]**.

The features used to distinguish between real and fake fingers are extracted from the image of the fingerprint. There are techniques such as in [1] and [10], in which the features used in the classifier are based on the specific fingerprint measurements, such as ridge strength, continuity and clarity. In contrast, some works use general feature extractors, such as Weber Local Descriptor (WLD) [11], which is a texture descriptor composed of differential excitation and orientation components. A local descriptor that uses local amplitude contrast (spatial domain) and phase (frequency domain) to form a bi-dimensional contrast-phase histogram was proposed in [12]. Both techniques achieve good results in standard benchmarks.

We approach the problem by experimenting two general feature extractors: Convolutional Networks (CN) that are, to the best of our knowledge, used for the first time for this task, and Local Binary Patterns (LBP), whose a multi-scale variant reported in [13] achieves good results in fingerprint liveness detection benchmarks. In opposition to more elaborated techniques that uses texture descriptors as features vectors, such as Local Phase Quantization (LPQ) [14], LBP with wavelets [15], and BSIF [16], our LBP implementation uses the original and uniform LBP coding schemes.

Moreover, we tested a variety of optional preprocessing techniques such as contrast normalization, frequency filtering, and region of interest (ROI) extraction. Augmented datasets [17] [18] are successfully used to increase the classifier's robustness against small variations by creating additional samples from image translations and horizontal reflections.

## II. METHODOLOGY

### A. Processing Flow

In this work, the pipelines used in training and testing can be broadly divided in four phases: preprocessing, feature extraction, dimensionality reduction, and classification.

An automatic and extensive search for the best combination of preprocessing operations, architectures and hyper-parameters was made during the 5x2 cross-validation phase [19], using the fastest computer configuration available from Amazon's Elastic Compute Cloud (EC2) services.

The implementation details of each phase will be explained in the following sub-sections.

### B. Preprocessing

Four preprocessing operations were carried out: image reduction, frequency filtering, region of interest (ROI) extraction and contrast equalization. The execution or non-execution of each operation in the final model is decided at validation time, that is, the combination of preprocessing operations that had the lowest validation error were included in the final model.

*Image Reduction*

In order to verify the effect of the spatial resolution in the classifier's performance, we apply image reduction using bilinear interpolation with different scales.

*Frequency Filtering*

We inspected how noise removal through a Gaussian low-pass filtering could improve results. We also tested the hypothesis that the relevant information to distinguish between false and real fingerprints is mostly in the high frequency components of the image by applying a Gaussian high-pass filter before extracting the features. The low-pass filter is implemented as the convolution of the input image by a Gaussian kernel and the high-pass filter is implemented as the subtraction of the original image by the low-pass filtered image. In our experiments, either high pass or low-pass filter was applied (never both) and the Gaussian kernels have a standard deviation of 3 pixels and size of 13x13 pixels.

*Region of Interest (ROI)*

As many fingerprints from some datasets, like Crossmatch sensor from LivDet 2013 competition, are not centered and the background represents a large part of the image, we created a simple ROI method using the following steps:

1. Apply morphological closing operation to highlight the region where the fingerprint lies. We used a box of size 21x21 as the structuring element, which is greater than the maximum ridges distances even in the largest images (that normally have greater ridge distances). This ensures that the fingerprint will become a continuous object after the operation.

2. Find the center of mass and the standard deviation of the image.

3. Get the region of interest: a rectangle centered in the center of mass, whose width and height are three times the standard deviations calculated in the previous step.

*Contrast Equalization*

We tested a technique called Contrast Limited Adaptive Histogram Equalization (CLAHE) [20], which is a variant of Adaptive Histogram Equalization (AHE) [21]. AHE computes several histograms, each corresponding to a distinct section of the image, and uses them to redistribute the lightness values of the image. It is, therefore, suitable for improving the local contrast of an image and bringing out more details. AHE has a tendency to overamplify noise in relatively homogeneous regions of an image. CLAHE prevents this by limiting the amplification by clipping the histogram at a predefined value before computing the neighborhood cumulative distribution function (CDF).

### C. Feature Extraction

Two different feature extractors were tested: Convolutional Networks (CN) with random weights and Local Binary Patterns (LBP).

*Convolutional Networks*

Convolutional Networks [22] are the state-of-the-art technique in a variety of image recognition benchmarks, such as MNIST [23], CIFAR-10 [23], CIFAR-100 [24], SVHN [23] and IMAGENET [18], and to the best of our knowledge, this is the first time it is employed in fingerprint liveness detection.

A classical convolutional network is composed of alternating layers of convolution and local pooling (i.e. subsampling). The aim of the convolutional layer is to extract patterns found within local regions of the inputted images that are common throughout the dataset, by convolving a template or filter over the inputted image pixels and outputting this as a feature map c, for each filter in the layer.

A non-linear function f(c) is then applied element-wise to each feature map c: a = f(c). A range of functions can be used for f(c), with tanh(c) and logistic functions being popular choices. In this paper we use a linear rectification f(c) = max (0; c) as the non-linearity function. In general, this has been shown [25] to have significant benefits over tanh() or logistic functions.

The resulting activations f(c) are then passed to the pooling layer. This aggregates the information within a set of small local regions, R, producing a pooled feature map s (normally of smaller size) as output. Denoting the aggregation function as pool(), for each feature map c we have:

$$s_j = \text{pool}(f(c_i)) \;\forall i \in R_j \quad (1)$$

where $R_j$ is the pooling region j in feature map c and i is the index of each element within it. Among the various types of pooling, max-pooling is commonly used, which selects the maximum value of the region $R_j$:

$$s_j = \max_{i \in R_j} a_i \quad (2)$$

The motivation behind pooling is that the activations in the pooled map *s* are less sensitive to the precise locations of structures within the image than the original feature map c. In a multi-layer model, the convolutional layers, which take the pooled maps as input, can thus extract features that are increasingly invariant to local transformations of the input image [26] [27]. This is important for classification tasks, since these transformations obfuscate the object identity. Achieving invariance to changes in position or lighting conditions, robustness to clutter, and compactness of representation, are all common goals of pooling.

Fig. 2 illustrates the feed-forward pass of a single layer convolution network. The input sample is convoluted with three random filters of size 5x5 (enlarged to make visualization easier), generating 3 convoluted images, which were then subject to non-linear function max(x,0), followed by a max-pooling operation and subsampled by a factor of 2.

Our convolutional networks use only random filters weights draw from a Gaussian distribution. Although the filter weights can be learned, as described in [28], filters with random weights can perform well and they have the advantage that they do not need to be learned [29] [30] [31].

It is a common practice to have a local contrast normalization layer (which is different from the Contrast Equalization previously described) between each convolution and pooling layer. The goal of this layer is to normalize pixels intensities based on its neighborhood. The operations of subtractive and divisive normalization, described below, are inspired by computational neuroscience models [32] [33] [34]. The subtractive normalization operation for a given 3D image patch $x_{ijk}$ is defined by:

$$V_{ijk} = x_{ijk} - \sum_{ipq} w_{pq} \cdot x_{i,j+p,k+q} \quad (3)$$

where $w_{pq}$ is a Gaussian weighting window normalized so that

$$\sum_{ipq} w_{pq} = 1 \quad (4)$$

i refers to the index of the third dimension of the image patch, j and k refer to the two dimensions of the image patch, p and q refer to the neighborhood region of the patch defined by j and k. The divisive normalization computes

$$y_{ijk} = v_{ijk}/\max(c, \sigma_{jk}) \quad (5)$$

where

$$\sigma_{jk} = \left(\sum_{ipq} w_{pq} \cdot v_{i,j+p,k+q}^2\right)^{1/2} \quad (6)$$

and c = 1 in our experiments.

*Local Binary Patterns*

Local Binary Patterns (LBP) are a local texture descriptor that have performed well in various computer vision applications, including texture classification and segmentation, image retrieval, surface inspection, and face detection [35]. The best current method for fingerprint liveness detection [13] uses this technique.

In its original version, the LBP operator assigns a label to every pixel of an image by thresholding each of the 8 neighbors of the 3x3-neighborhood with the center pixel value and considering the result as a unique 8-bit code representing the 256 possible neighborhood combinations. As the comparison with the neighborhood is done with the central pixel, the LBP is an illumination invariant descriptor. The operator can be extended to use neighborhoods of different sizes [36].

Another extension to the original operator is the definition of so-called *uniform patterns*, which can be used to reduce the length of the feature vector and implement a simple rotation-invariant descriptor [36]. An LBP is called uniform if the binary pattern contains at most two bitwise transitions from 0 to 1 or vice versa when the bit pattern is considered circular.

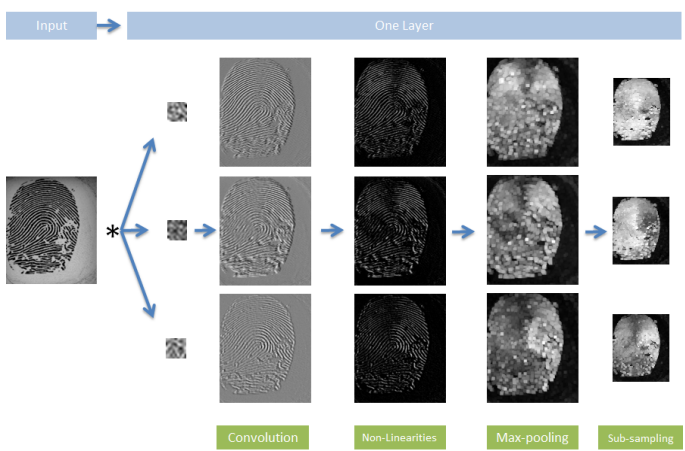

Fig. 2   Illustration of a sequence of operations performed by a single layer convolutional network in a sample image.

The number of different labels of LBP is reduced from 256 to just 10 in the uniform pattern.

The normalized histogram of the LBPs (with 256 and 10 bins for non-uniform and uniform operators, respectively) is used as a feature vector. The assumption underlying the computation of a histogram is that the distribution of patterns matters, but the exact spatial location does not. Thus, the advantage of extracting the histogram is spatial invariance property. To investigate if location matters to our problem, we also implemented the method presented in [37], for face recognition, where the LBP filtered images are equally divided in rectangles and their histograms are concatenated to form a final feature vector.

### D. Feature Normalization, Dimensionality Reduction and Whitening

After the feature extraction phase, each dimension of the dataset is independently normalized to zero mean and unit variance. This is normally required because many elements used in the objective function of a learning algorithm (such as the RBF kernel of Support Vector Machines) assume that all features are centered on zero and have variance in the same order. If a feature has a variance that is orders of magnitude larger than others, it might dominate the objective function and make the estimator unable to learn from other features correctly as expected.

The normalized data is then subject to dimension reduction by using Randomized Principal Components Analysis [38] [39], which is faster than original Principal Components Analysis (PCA) because it limits computation to an approximated estimation of the singular vectors that will actually perform the transformation.

Whitening transformation [40] [41], also called Sphering, is applied after PCA to normalize the variances of the principal components by dividing them by their standard deviations, which has been shown to improve results in computer vision classification tasks [42]. Denoting the PCA rotated components by $y_i$, this means we compute

$$s_i = \frac{y_i}{\sqrt{var(y_i)}} \quad (7)$$

to get whitened components $s_i$. This is often useful if the classification model makes assumptions on the isotropy of the signal, which is the case for Support Vector Machines with the RBF kernel.

### E. Classification

As the final step of the pipeline, a Support Vector Machines (SVM) with a Gaussian Radial Basis Function (RBF) kernel is used as they have shown slightly better performance than the linear kernel. More precisely, preliminary experiments with the linear kernel show that there is a decrease of 0.5-1% in the accuracy, on average.

One might question the use of PCA before SVM, as the latter is supposed to handle high-dimensionality data already. We performed other experiments, not reported in this paper, without PCA and there is an average decrease of 2% in the accuracy. Our architectural choice is also based on the results published by [43], which shows that there is an improvement when PCA is applied before SVM, and by [44], which shows that PCA before SVM can be used to speed-up training time with no major impact in the accuracy.

### F. Increasing Classifier's Generalization through Dataset Augmentation

Dataset Augmentation is a technique that consists in artificially creating slightly modified samples from the original ones. Using them during training, it is expected that the classifier will become more robust against small variations that may be present in the data, forcing it to learn larger (and possible more important) structures. It has been successfully used in computer vision benchmarks such as in [45], [46], and [47].

Our dataset augmentation implementation is similar to the one presented in [18]: from each image of the dataset five smaller images with 80% of each dimension of the original images are extracted: four patches from each corner and one at the center. For each patch, horizontal reflections are created. As a result, we obtain a dataset that is 10 times larger than the original one: 5 times are due to translations and 2 times are due to reflections. At test time, the classifier makes a prediction by averaging the individual predictions on the ten patches.

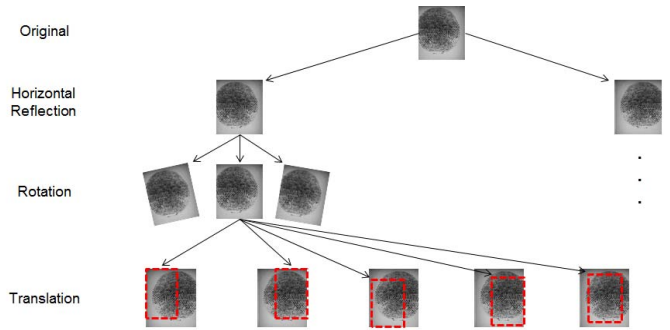

Fig. 3 Illustration of three types of transformations for dataset augmentation: Horizontal Reflections, Rotations, and Translations.

### G. Performance Metrics

The classification results were evaluated by the Average Classification Error (ACE), which is the standard metric for evaluation the LivDet competitions. It is defined as

$$ACE = (FPR + FNR)/2 \quad (8)$$

where *FPR* (False Positive Rate) is the percentage of misclassified live fingerprints and *FNR* (False Negative Rate) is the percentage of misclassified fake fingerprints.

### H. Implementation Details

The algorithms were implemented in Python using build-in functions from Numpy, Scipy, Scikit-Image and Scikit-Learn packages, except for the Convolutional Networks, which we used an efficient package from [48], and Cross-Validation/Grid-Search, that we wrote our own code using Numpy.

We wrote an improved cross-validation/grid-search algorithm for choosing the best combination of hyper-parameters, in which each element of pipeline is computed only when its training data is changed (the term "element" refers to operations such as preprocessing, feature extraction, dimensionality reduction or classification). This modification speeded-up the validation phase in approximately 10 times, although the gain can greatly vary as it depends on the element types and number of hyper-parameters chosen.

An important aspect of this work is that the algorithms were run on cloud service computers, where the user can rent virtual computers and pay only for the hours that the machines are running. To train the algorithms, we used the fastest Amazon EC2 HPC instance available, with 32 cores and 60 GB of RAM that allowed us to run dataset augmented experiments in a few hours – otherwise it would take weeks to perform them.

## III. EXPERIMENTS

### A. Datasets

We used the datasets provided by the Liveness Detection Competition (LivDet) in the years of 2009 [9], 2011 [49] and 2013 [50].

LivDet 2009 comprises almost 18,000 images from real and fake fingerprints acquired from three different sensors (Biometrika FX2000, Crossmatch Verifier 300 LC, and Identix DFR 2100). Fake fingerprints were obtained from three different materials: Gelatin, Play Doh, and Silicone. Approximately one third of the images of the dataset are used for training and the remaining for testing.

LivDet 2011 comprises 16,000 images acquired from four different sensors (Biometrika FX2000, Digital 4000B, Italdata ET10, and Sagem MSO300), each having 2000 images from fake and real fingerprints. Half of the dataset is used for training and the other half for testing. Fake fingerprints were obtained from four different materials: Gelatin, Wood Glue, Eco Flex, and Silgum.

LivDet 2013 comprises 16,000 images acquired from four different sensors (Biometrika FX2000, Crossmatch L SCAN GUARDIAN, Italdata ET10, and Swipe), each having approximately 2,000 images from fake and real fingerprints. Half of the dataset is used for training and the other half for testing. Fake fingerprints were obtained from five different materials: Gelatin, Latex, Eco Flex, Wood Glue, and Modasil.

In all datasets, the real/fake fingerprint ratio is 1/1 and they are equally distributed between training and testing sets. The sizes of the images vary from sensor to sensor, ranging from 240x320 to 700x800 pixels.

### B. Pipelines

TABLE I lists the pipelines used in the experiments. The preprocessing step is omitted but it was executed in all pipelines. Due to the large number of hyper-parameters that defines a pipeline (like number of layers, filter sizes in each layer, number of principal components, etc) it was not possible to display them for all of the 11 datasets. Hence, the full list of hyper-parameters searched and selected, and the corresponding scores obtained in the validation phase for each dataset can be found in http://adessowiki.fee.unicamp.br/adesso/wiki/Demo/fingerprint/view/.

TABLE I. SUMMARY OF THE PIPELINES EVALUATED IN THIS WORK

| Pipeline | Description |
|---|---|
| CN + PCA + SVM | Features are extracted using Convolutional Networks. The feature vector is reduced using PCA and then fed into a SVM classifier using (Gaussian) RBF kernel. |
| LBP + PCA + SVM | Features are extracted using LBP. The feature vector is reduced using PCA and then fed into a SVM classifier with (Gaussian) RBF kernel. |
| AUG + LBP + PCA + SVM | Dataset is artificially augmented, and the pipeline follows as in the LBP+PCA+SVM pipeline. |
| AUG + CN + PCA + SVM | Dataset is artificially augmented, and the pipeline follows as in the CN+PCA+SVM pipeline. |

## IV. RESULTS

The average error for each testing dataset is shown on TABLE II. The state-of-the-art results for LivDet 2009, 2011 and 2013 datasets were taken from [11], [12], and [50], respectively. It is important to mention that [50] was a participant of the LivDet2013 competition and it did not have access to the testing datasets during the development phase. The results from the other two state-of-the-art techniques, [11] and [12], are post-competition results, that is, the authors had access to testing datasets during the development phase.

TABLE II. AVERAGE CLASSIFICATION ERROR ON TESTING DATASETS

| Technique | | State of the art | Aug LBP PCA SVM | LBP PCA SVM | Aug CN PCA SVM | CN PCA SVM |
|---|---|---|---|---|---|---|
| LivDet 2013 | Crossmatch | 31.2 [50] | 49.45 | 49.87 | **3.29** | 5.2 |
| | Swipe | 14.07 [50] | **3.34** | 4.02 | 7.67 | 5.97 |
| | Italdata | 3.5 [50] | **2.3** | 55.45 | 2.45 | 47.65 |
| | Biometrika | 4.7 [50] | 1.7 | 25.65 | **0.8** | 4.55 |
| LivDet 2011 | Italdata | 11.0 [12] | 12.34 | 23.68 | **9.27** | 5.09 |
| | Biometrika | **4.9** [12] | 8.85 | 8.2 | 8.25 | 9.9 |
| | Digital | 4.2 [12] | 4.15 | 3.85 | 3.65 | **1.9** |
| | Sagem | **2.7** [12] | 7.54 | 5.56 | 4.64 | 7.86 |
| Livdet 2009 | Biometrika | **0.31** [11] | 10.44 | 50 | 9.23 | 9.49 |
| | Crossmatch | 3.13 [11] | 3.65 | 6.81 | **1.78** | 3.76 |
| | Identix | 1.16 [11] | 2.64 | 0.95 | **0.8** | 2.82 |
| | Average | 7.35 | 9.67 | 21.28 | **4.71** | 9.47 |

The LBP without dataset augmentation pipelines seems to suffer from overfitting, since it has a very low cross-validation error (close to 0%) whereas a large error rate from the testing datasets of Crossmatch 2013, Italdata 2013, Biometrika 2013, Italdata 2011, and Biometrika 2009. However, when dataset augmentation is used, the cross-validation error increases but the testing error decreases, except for Crossmatch 2013 dataset, which is discussed later. This is a good indication that dataset augmentation can be used to prevent overfitting.

The results for Crossmatch 2013 dataset using LBP presents error rates close to zero at validation time and around 50% at test time, even when using augmented datasets. It can be noticed from LivDet 2013 competition results that this dataset is particularly difficult to generalize, since nine of the eleven participants presented error rates greater than 45%. CN performs very well (3.28%), which suggests that the problem occurs mostly when extracting features with LBP.

Overfitting seems not to be a problem when using Convolutional Networks, except for the Italdata 2013 dataset, which has a validation error of 0.15% and a testing error of 47.65%. On overall, CN without dataset augmentation have a similar performance to LBP with dataset augmentation.

When using augmented dataset with convolutional networks, we achieved a test error rate of 4.75% (averaged from all datasets), which represents a reduction of 35% when compared to the best previously published results (7.35% error, on average).

The optimum number of layers in the Convolutional Networks depends on the dataset: it varies from two to five layers. The fact that one layer networks were not selected confirms that the deep architectures perform better on the task than the shallow ones. On overall, the best convolution shapes are 9x9 for the first layers and 5x5 for the last layers. The best pooling shapes are 7x7 for the first layers and 5x5 for the last layers. The best quantity of filters was 256 or 512 for the first layers and 1024 or 2048 for the last layers. We could not find a relation between architectures and dataset characteristics, such as image size and foreground/background ratio to explain the choices for the best parameters. Similarly, the best LBP operator (uniform or non-uniform) and number of divisions depend on the dataset.

On average, the PCA models selected in the validation phase reduced the input vectors to 20% of their original dimensions, which represents a variance close to 100%. This is an indication that the vectors extracted using either CN or LBP still contain redundant information and reducing the dimensions using PCA can be advantageous. This statement is confirmed by preliminary empirical results not shown in this paper: pipelines that use PCA have greater accuracy rates than the ones that do not use it.

For the majority of datasets and models, preprocessing operations (contrast equalization, ROI, etc) did not improve accuracy. For contrast equalization this is not surprising, since both LBP and CN offer some invariance to illumination differences. Regarding to ROI extraction, the backgrounds are mostly composed of white pixels, which results in components in the final extracted vector that have low variances and are probably discarded by the dimensionality reduction and the SVM classifier during the training phase. In addition, the histogram extraction in the LBP pipeline and the pooling/sub-sampling operation in the CN offer translation invariance.

The hypothesis that the relevant information for liveness detection lies either in the low-frequency or in the high-frequency components of the image was not confirmed. Both low-pass and high-pass filtering decreased accuracy during validation, which suggests that the structures that differentiate false from real fingerprints do not have exclusively low or high frequency components.

As 50 % image reduction improved performance on five datasets (Italdata 2011, Italdata 2013, Biometrika 2011, Biometrika 2013 and Identix 2009) when using LBP, and only in one dataset (Identix 2009) when using CN. Based on these differences, we conclude that there is not an optimal image size for classification; it depends not only on the sensor type but also on the acquisition dataset and the transformations used.

In real applications, a good fingerprint liveness detection system must be able to classify the images in a short amount of time. On average, AUG+LBP+PCA+SVM and AUG+CN+PCA+SVM pipelines take less than 300 ms and 600 ms, respectively, to classify a single image on a regular single core computer (1.8 GHz, 64-bit, with 2 GB memory), meaning that the algorithms are able to run in any conventional low-end PC. The training time for each model for LBP and CN are around 20 minutes and 1.5 hours, respectively, on the Amazon's EC2 machine.

## V. CONCLUSIONS

Two SVM classification models were tested for fingerprint liveness detection. One based on the Local Binary Patterns (LBP) and the other based on Convolutional Networks (CN). The CN presented the best performance (4.71%, on average), but they are slower to train and more complex to design than LBP. Compared to the state-of-the art techniques, CN has the best performance in 5 of the 11 datasets, while our LBP technique has the best performance in 2 of the 11 datasets.

Preprocessing operations such as region of interest extraction and Contrast Equalization did not help to improve accuracy, mainly because the feature extractors already offer some robustness against illumination and translation variances. PCA and Whitening are necessary, since the data has redundant dimensions after the feature extraction phase.

Dataset augmentation demonstrated to play an important role to increase accuracy and it is simple to implement. We claim that the method should always be considered if one has enough computational power.

We believe that the main contributors for the good achieved results were the large datasets we used, like images in their original sizes, augmented datasets, and the large number of layers and filters in the convolutional networks. With faster computers, we could execute a large number of experiments due to faster training/validation iteration. The emerging high power cloud computing platforms make the building of increasingly large experiments affordable by renting ready-to-run virtual computer infrastructure.

## VI. FUTURE WORK

Further experiments will include learning the filters' weights in the convolutional networks, as [30] reported that a better performance is achieved when the network is trained.

Given the promising results provided by the dataset augmentation, more types of image transformations should be included, such as artificially creating images with uneven illumination and with random noise. We want to know the limits of the technique: how many times can the dataset be artificially augmented with an improvement in performance? Also, training one classifier per transformation type, as implemented in [47], may lead to better results.